\documentclass[letterpaper, 10 pt, conference]{ieeeconf}

\IEEEoverridecommandlockouts                              

\usepackage{amsmath} 
\usepackage{cite}
\usepackage{graphicx}

\title{\LARGE \bf
Passive Phase-Oriented Impedance Shaping for Rapid Acceleration in Soft Robotic Swimmers
}

\author{Qimin Feng$^{1,*,\dagger}$, Orion A. Roberts$^{1,\dagger}$ and Qiang Zhong$^{1}$
\thanks{$^{1}$Department of Mechanical Engineering, Iowa State University, Ames, IA 50011, USA. $^{*}$Correspondence: Qimin Feng ({\tt\small qmfeng11@iastate.edu}). $^{\dagger}$Equal contribution.}%
}

\begin{document}
\maketitle
\thispagestyle{empty}
\pagestyle{empty}

\begin{abstract}
Rapid acceleration and burst maneuvers in underwater robots depend less on maintaining precise resonance and more on force--velocity phase alignment during thrust generation. In this work, we investigate constrained-layer damping (CLD) as a passive mechanism for frequency-selective impedance shaping in soft robotic swimmers. Unlike conventional stiffness-tuning approaches, CLD selectively amplifies the dissipative component of bending impedance while preserving storage stiffness, passively shifting the impedance composition toward dissipative dominance as actuation frequency increases. We characterize this behavior through dry impedance measurements, demonstrate that CLD enhances thrust and alters force--motion phase relationships across Strouhal numbers in constrained propulsion tests, and validate that passive impedance shaping yields a nearly five-fold increase in peak acceleration and a three-fold increase in terminal velocity in unconstrained swimming trials. These results establish phase-oriented passive impedance modulation as a simple, control-free pathway for improving transient propulsion in soft robotic systems.
\end{abstract}


\section{Introduction}
Underwater robots must frequently execute rapid acceleration maneuvers—burst starts, predator-escape responses, station-keeping under gusts, and obstacle avoidance—that demand large thrust impulses delivered within only a few flapping cycles. In biological swimmers, such transient performance is achieved through a sophisticated interplay of active muscle contraction and passive viscoelastic tissue properties, enabling robust propulsion across rapidly changing flow conditions~\cite{long1996importance}. Inspired by these observations, a large body of work in bio-inspired robotics has sought to exploit structural flexibility and resonance tuning to improve propulsive performance~\cite{nashif1991vibration,mead1969forced,ross1959damping,zhong2021tunable,quinn2022tunable}.  
However, nearly all existing stiffness-tuning strategies have been developed and evaluated under steady-state cruising conditions, leaving transient acceleration regimes largely unaddressed.

A fundamental reason to look beyond resonance for rapid acceleration is that the physical conditions required for resonance cannot be maintained during transient motion. As a swimmer accelerates, its forward velocity changes continuously, causing the reduced frequency, added mass, and fluid damping to evolve on a cycle-by-cycle basis. Because resonance peaks are inherently narrow and sensitive to these parameters, the system is driven away from its resonant condition almost immediately after acceleration begins. More importantly, even if dynamic resonance could be preserved, it would not necessarily maximize the quantity that governs acceleration: the net thrust impulse per cycle is maximized when the hydrodynamic force is phase-aligned with the propulsor velocity; any phase mismatch diverts actuation energy into lateral fluid motion or non-propulsive elastic storage. Force--velocity phase alignment, rather than resonance amplitude, therefore plays the dominant role in transient propulsion performance~\cite{anderson1998oscillating}.

\begin{figure}[t]
  \centering
  \includegraphics[width=\columnwidth]{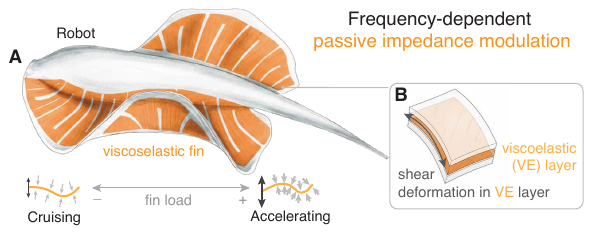}
  \caption{\textbf{Frequency-dependent passive impedance modulation in a viscoelastic soft fin.}
(A) A soft robot equipped with a viscoelastic fin can passively modulate its effective mechanical impedance in response to frequency-varying hydrodynamic loading, reducing reliance on precise active stiffness control.
(B) The fin incorporates a constrained-layer damping (CLD) element, where shear deformation in the viscoelastic (VE) layer increases with loading frequency, resulting in a shift of the effective impedance composition toward increased damping at higher frequencies.}
  \label{fig:1}
  \vspace{-1.5em}
\end{figure}

Fluid--structure interaction further complicates this picture. As the oscillation frequency increases (i.e., at higher Strouhal numbers), the hydrodynamic loading on the propulsor grows, and the added mass and fluid damping modify the effective impedance of the oscillating structure~\cite{jing2012effects,shah2024controlling}~(Fig.~\ref{fig:1}(A)). If the structural impedance does not adapt to these changing loads, the phase relationship between force and motion progressively deteriorates, leading to reduced thrust and inefficient energy transfer. This suggests that high-acceleration propulsion requires a mechanism capable of reshaping the effective impedance composition—specifically, the relative contributions of elastic (storage) and dissipative (loss) components—in a frequency-dependent manner that maintains favorable phase alignment as operating conditions evolve.

Active stiffness modulation has been proposed to regulate this interaction in robotic platforms. Approaches include impedance control with force feedback~\cite{xu2021high}, hybrid actuation combining compliant and rigid elements~\cite{bai2025waypoint}, and high-bandwidth closed-loop stiffness 
regulation~\cite{soto2025improving}. Although effective in principle, these strategies face significant practical limitations. In multi-degree-of-freedom soft structures—such as multi-joint robotic fish or distributed flexible fins—each degree of freedom requires an independent sensing and actuation channel for spatially coordinated stiffness tuning, leading to prohibitive mechanical complexity. Moreover, variable-stiffness smart materials (e.g., shape-memory alloys or polymers) typically exhibit response times far slower than the actuation frequencies relevant to swimming ($\sim$1--5~Hz), precluding real-time impedance adaptation. Furthermore, force sensing in underwater environments is inherently noisy, and the high-bandwidth feedback loops required for impedance control add cost, weight, and failure modes that are particularly problematic in miniaturized or untethered soft robotic systems. 
These limitations motivate the search for passive structural mechanisms that can shape effective impedance and force timing without relying on active regulation—embedding ``mechanical intelligence'' directly within the morphology of the propulsor~\cite{laschi2016soft}.

In this work, we investigate constrained-layer damping (CLD)---a well-established vibration mitigation architecture in aerospace and structural engineering~ \cite{ross1959damping,mead1969forced,johnson1982finite,nashif1991vibration,alvelid2007modelling,allen2010analyticity}---as a passive mechanism 
for frequency-selective impedance shaping in soft robotic swimmers (Fig.~\ref{fig:1}). Unlike conventional stiffness-tuning approaches that primarily modify elastic rigidity, CLD selectively amplifies the dissipative (loss) component of bending impedance while maintaining nearly constant storage stiffness. As the actuation frequency increases during acceleration, the viscoelastic core of the CLD structure undergoes increased shear deformation, progressively shifting the impedance composition from elastic-dominant toward dissipative-dominant (Fig.~\ref{fig:1}(B)). We demonstrate that this passive, frequency-dependent shift in impedance composition drives the force--velocity phase relationship toward alignment, thereby enhancing thrust impulse and rapid acceleration performance without active stiffness control.

The principal contributions of this work are:
\begin{enumerate}
    \item We propose CLD as a passive, frequency-selective impedance shaping 
    mechanism for soft robotic propulsion and characterize its complex bending 
    impedance under controlled conditions.
    \item We demonstrate that CLD systematically enhances thrust generation and 
    alters force--motion phase relationships across Strouhal numbers, with PIV 
    confirming improved vortex dynamics.
    \item We validate that passive impedance shaping yields a nearly five-fold 
    increase in peak acceleration in unconstrained swimming trials, establishing 
    a simple, control-free pathway for transient propulsion enhancement.
\end{enumerate}

\section{CLD Modeling and Fabrication}

\begin{figure}[h]
  \centering
  \includegraphics[width=\columnwidth]{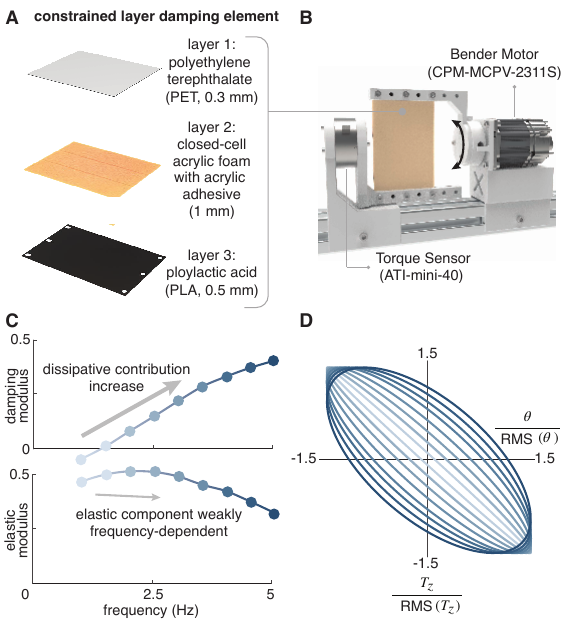}
  \caption{\textbf{CLD module fabrication and frequency-dependent impedance characterization.}
  (A)~Exploded view of the symmetric CLD sandwich structure: a PLA base plate 
  (0.5~mm) laminated with a closed-cell acrylic foam viscoelastic core (1~mm) 
  and a PET constraining layer (0.3~mm) on each side.
  (B)~Bender test apparatus. A Teknic ClearPath motor prescribes sinusoidal 
  angular oscillations while an ATI Mini-40 sensor records the resistive torque.
  (C)~Storage modulus $K'$ and loss modulus $K''$ versus actuation frequency. 
  $K''$ increases with frequency, indicating growing dissipative contribution, 
  while $K'$ remains approximately constant.
  (D)~Normalized torque--angle phase portraits. Increasing actuation frequency 
  (light to dark) expands the enclosed loop area, confirming higher energy 
  dissipation per cycle.}
 
  \label{fig:2}
  \vspace{-1.5em}
\end{figure}

\subsection{CLD Mechanical Model}
We consider a symmetric constrained-layer-damped (CLD) sandwich beam comprising a structural base plate, a viscoelastic (VE) core, and a thin constraining layer on each side (Fig.~\ref{fig:2}(A)). When the beam undergoes bending, the constraining layers restrict free deformation of the VE core, converting a portion of the flexural motion into through-thickness shear within the core. Because the VE material is rate-dependent, the resulting shear stress—and therefore the energy dissipated per cycle—increases with oscillation frequency. This mechanism gives rise to a complex, frequency-dependent bending stiffness

\begin{equation}
    K^{*}(\omega) = K'(\omega) + i\,K''(\omega),
\end{equation}

where the storage component $K'(\omega)$ represents elastic energy return and the loss component $K''(\omega)$ represents dissipative energy transfer per cycle. In the low-frequency range relevant to aquatic propulsion ($0$--$5$~Hz), classical sandwich-beam theory~\cite{ross1959damping, mead1969forced} predicts that $K'$ remains approximately constant—set primarily by the base-plate stiffness—while $K''$ grows with frequency as the shear strain rate in the VE core increases. The net effect is a progressive shift in impedance composition from elastic-dominant toward dissipative-dominant with increasing actuation frequency. Crucially, this mechanism is distinct from simply increasing 
structural rigidity. In a CLD structure, the viscoelastic core ensures that the increase is concentrated in $K''$, selectively amplifying dissipation while leaving the elastic response largely 
unchanged.

\subsection{Impedance Extraction}

To extract the complex bending stiffness from experimental data, we apply a single-frequency harmonic regression (lock-in method) to the measured angular displacement $\theta(t)$ and resistive torque $T_z(t)$. At each actuation frequency $\omega$, we decompose the time-domain signals into their fundamental Fourier components, yielding complex amplitudes $\hat{\theta}$ and $\hat{T}$. The complex stiffness follows as
\begin{equation}
    K^*(\omega) = \frac{\hat{T}}{\hat{\theta}}\,,
\end{equation}
with the storage modulus $K' = \Re\{K^*\}$ and loss modulus $K'' = \Im\{K^*\}$ representing the in-phase (elastic) and quadrature (dissipative) components of the structural response, 
respectively.

To track how the impedance composition evolves with frequency, we define the elastic and dissipative fractions as
\begin{equation}
    f_{\text{elastic}} = \frac{K'}{K' + K''},
    \qquad
    f_{\text{dissipative}} = \frac{K''}{K' + K''}\,,
\end{equation}
where both $K'$ and $K''$ are non-negative over the tested frequency range. These metrics provide a direct, dimensionless measure of the relative contributions of energy storage and dissipation to the 
overall bending impedance at each operating point.

\subsection{Fabrication of Soft CLD Modules}
Soft CLD modules are fabricated by symmetrically laminating a polylactic acid (PLA, $0.5$~mm) base plate with a closed-cell acrylic foam viscoelastic core ($1$~mm, acrylic adhesive) and a polyethylene 
terephthalate (PET, $0.3$~mm) constraining layer on each side, following the layer numbering in Fig.~\ref{fig:2}(A). The assembled module is coupled to an ATI Mini-40 force/torque sensor via a rigid 
adapter for torque measurement. Module geometry—including length (100 mm), width (76.5 mm), and layer thicknesses—is selected to place the primary bending modes within the $0$--$5$~Hz actuation range while maintaining ease of fabrication and repeatable assembly.

\section{Structural Impedance Characterization}

\subsection{Dry-Test Setup}

The frequency-dependent impedance of the CLD module was characterized using the bender apparatus shown in Fig.~\ref{fig:2}(B). A Teknic ClearPath motor prescribes sinusoidal angular oscillations at 
frequencies from $0$ to $5$~Hz with a peak-to-peak amplitude of $18^\circ$ ($\pm9^\circ$ from equilibrium), while an inline ATI Mini-40 force/torque sensor records the resistive torque. All tests were performed with 100\% CLD coverage of the base PLA plate. Data were acquired at $200$~Hz using a National Instruments DAQ card, providing at least 40 samples per cycle at the highest test frequency.

\begin{figure}[!th]
  \centering
  \includegraphics[width=\columnwidth]{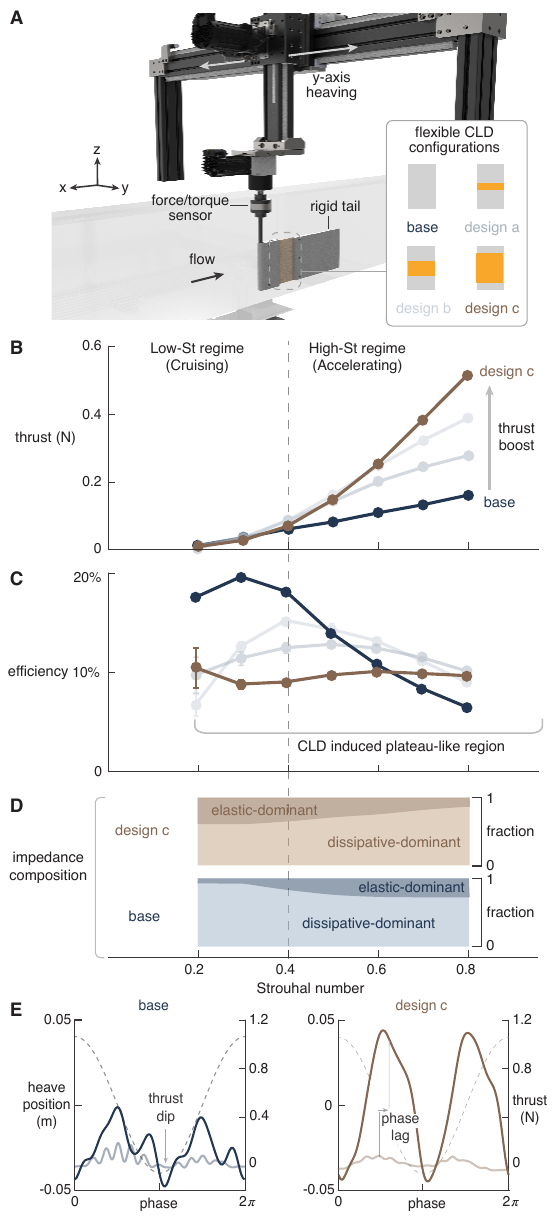}
  \caption{\textbf{Constrained steady-state propulsion performance.}
  (A)~Heaving test setup with four CLD configurations (0\%, 16.7\%, 
  33.3\%, 66.7\% coverage).
  (B)~Mean thrust versus $St$; shaded regions indicate $\pm 1$ standard 
  deviation over three trials.
  (C)~Propulsive efficiency versus $St$; error bars as in (B).
  (D)~Effective impedance composition versus $St$. The baseline remains 
  dissipative-dominant; design~c shifts progressively from elastic-dominant 
  toward dissipative dominance.
  (E)~Instantaneous thrust and heave position over one cycle at $St=0.8$.}
  \label{fig:3}
  \vspace{-1.5em}
\end{figure}

\subsection{Results}

The measured impedance confirms the frequency-selective dissipation predicted by the CLD sandwich-beam model (Section~II-A). As shown in Fig.~\ref{fig:2}(C), the storage modulus $K'$ remains approximately 
constant across the tested frequency range, indicating that the elastic energy-return capability of the module is preserved independent of actuation frequency. In contrast, the loss modulus $K''$ increases monotonically with frequency, reflecting the growing shear dissipation in the viscoelastic core.

This trend is corroborated by the normalized torque--angle phase portraits (Fig.~\ref{fig:2}(D)). As frequency increases (light to dark), the hysteresis loops widen progressively, with the enclosed area—proportional to dissipated energy per cycle—expanding substantially. The transition from narrow, nearly linear trajectories at low frequencies to broad ellipses at high frequencies provides direct visual confirmation that the CLD mechanism shifts the impedance composition toward dissipative dominance with increasing actuation rate.

\section{Constrained Steady-state Propulsion Analysis}

\subsection{Experimental Setup}

Constrained steady-state propulsion tests were conducted using a four-axis Cyber-Physical System (CPS) in a recirculating flow facility. Each CLD module was mounted to an ATI Mini-40 force/torque sensor via a 7~mm carbon-fiber rod, with a rigid teardrop-shaped tail attached to the trailing edge to form the propulsive hydrofoil (Fig.~\ref{fig:3}(A)). The foil was held at a fixed streamwise position while the CPS $y$-axis motor prescribed sinusoidal heave motion with an amplitude of $0.08$~m ($\pm0.04$~m from equilibrium) at frequencies from $0.5$ to $2$~Hz in $0.25$~Hz increments, perpendicular to a freestream velocity of $0.2$~m/s along $x$. This combination yields a Strouhal number range of $St=0.2$--$0.8$, spanning cruising through rapid-acceleration regimes typical of aquatic swimmers. Four CLD configurations were tested: baseline (0\% coverage), 16.7\% (design~a), 33.3\% (design~b), and 66.7\% (design~c). Each condition was repeated three times and cycle-averaged. Hydrodynamic loads were sampled at $200$~Hz; heave commands were issued at $4000$~Hz via an NI DAQ system running a Simulink real-time controller.

Phase-resolved particle image velocimetry (PIV) was performed to characterize the propulsive wake. A continuous laser sheet illuminated seeding particles in the mid-span plane, and images were captured with a Photron Nova R3 high-speed camera. Vector fields were computed using DaVis (LaVision) \cite{feng2025lift}.

\subsection{Thrust and Efficiency}

Mean thrust and propulsive efficiency across the $St$ range are shown in Figs.~\ref{fig:3}(B) and~\ref{fig:3}(C). In the low-$St$ cruising regime ($St=0.2$--$0.4$), all four configurations produce 
comparable thrust. Above $St=0.4$, however, thrust diverges sharply: at $St=0.8$, design~c generates $0.51$~N compared with $0.16$~N for the baseline—an increase exceeding 200\%. Designs~a 
and~b scale continuously between these extremes, confirming that the propulsive benefit increases monotonically with CLD coverage.

Efficiency trends reveal a complementary trade-off (Fig.~\ref{fig:3}(C)). The baseline peaks at approximately 20\% near $St=0.3$ but declines rapidly at higher $St$, falling below 
design~c for $St > 0.7$. Design~c, by contrast, maintains a plateau-like efficiency of roughly 10\% across the entire $St$ range, indicating that the CLD-induced impedance shift enables the foil to 
sustain useful thrust production without the sharp efficiency collapse seen in the baseline. Designs~a and~b exhibit transitional profiles between these two limiting behaviors.

\subsection{Impedance Composition and Phase Analysis}

The effective impedance composition of the coupled fluid--structure system (Fig.~\ref{fig:3}(D)) provides a mechanistic explanation for the divergent thrust trends. The baseline remains overwhelmingly dissipative-dominant across all tested $St$, with its elastic fraction widening only marginally at high frequencies. Design~c, in contrast, begins with a substantial elastic-dominant fraction 
(${\sim}40\%$) at $St=0.2$ and shifts progressively toward dissipative dominance as $St$ approaches $0.8$. This continuous, passive rebalancing of impedance composition mirrors the frequency-dependent loss-modulus growth observed in the dry characterization (Section~III), confirming that the structural-level tuning mechanism is preserved under hydrodynamic loading. This distinction is critical: if CLD merely increased elastic stiffness, the impedance composition would remain constant across 
$St$ and no phase shift would occur. The progressive migration toward dissipative dominance confirms that CLD reshapes the impedance composition rather than uniformly scaling it.

The consequences for force timing are visible in the instantaneous thrust traces (Fig.~\ref{fig:3}(E)). At $St=0.8$, the baseline produces a highly irregular thrust waveform with multiple dips and erratic peaks, symptomatic of poor phase coordination between structural motion and fluid forcing. Design~c, by contrast, outputs a smooth waveform with two distinct thrust peaks per cycle and a clear phase lag relative to the heave input, consistent with the hypothesis that increased structural dissipation shifts force production toward velocity-aligned timing. This provides a passive structural route to phase modulation, complementing prior foil studies based on prescribed heave--pitch phasing~\cite{vanburen2019scaling,floryan2017scaling}.

\begin{figure}[b]
  \vspace{-1.5em}
  \centering
  \includegraphics[width=\columnwidth]{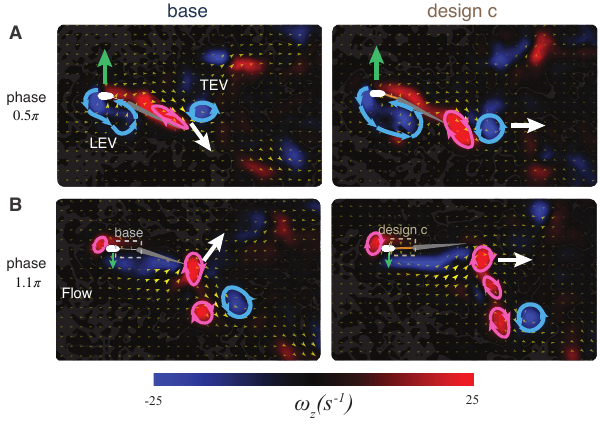}
  \vspace{-1.5em}
  \caption{\textbf{Propulsive wake evolution.} (A) Phase-resolved PIV flow field at $0.5\pi$ for the baseline and design~c, highlighting LEV/TEV structures. (B) Flow field at $1.1\pi$ during heaving reversal, showing vortex development along the foil chord.}
  \label{fig:4}
\end{figure}

\begin{figure*}[!t]
  \centering
  \vspace*{0.5mm}
  \includegraphics[width=\textwidth]{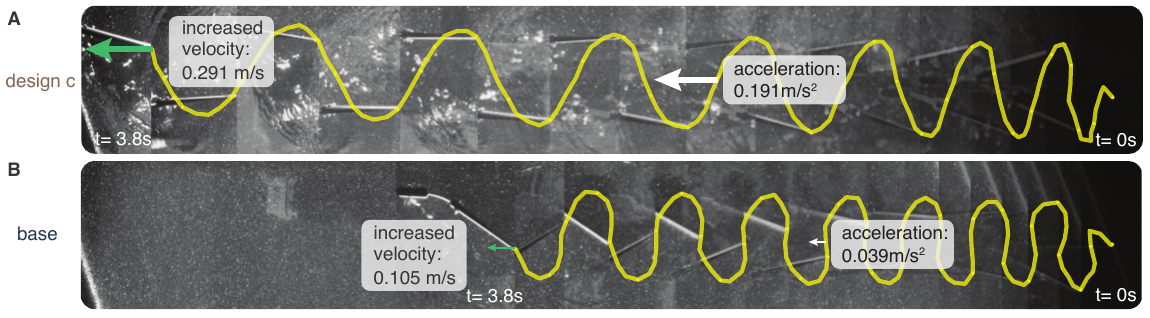}
  \vspace{-1.5em}
  \caption{\textbf{Unconstrained rapid acceleration at $St=0.8$.}
  (A)~Design~c achieves a peak acceleration of $0.191~\mathrm{m/s^2}$ and 
  terminal velocity of $0.291$~m/s over a $3.8$~s trial.
  (B)~The baseline reaches $0.039~\mathrm{m/s^2}$ acceleration and 
  $0.105$~m/s terminal velocity over the same duration. Yellow traces 
  show the trailing-edge trajectory.}
  \label{fig:5}
  \vspace{-1.5em}
\end{figure*}

\subsection{Wake Dynamics}

Phase-resolved PIV corroborates the force-level observations and provides a direct flow-physics explanation for the thrust waveforms in Fig.~\ref{fig:3}(E). At a phase of $0.5\pi$ (Fig.~\ref{fig:4}(A))—corresponding to the peak thrust region in design~c—the CLD-enhanced foil forms a large, coherent leading-edge vortex (LEV) that remains firmly attached, and its trailing-edge vortex (TEV) is shed in the streamwise direction. This well-organized vortex system is consistent with the smooth, high-amplitude thrust peak that design~c produces near $0.5\pi$ at $St=0.8$. The baseline, by contrast, generates a smaller, weaker LEV with a laterally deflected TEV at the same phase, consistent with the thrust dip visible in its waveform near $0.5\pi$.

The contrast sharpens at the heaving reversal ($1.1\pi$, Fig.~\ref{fig:4}(B)). The baseline retains its previous pitch angle past the motion peak, delaying LEV initiation and producing disorganized, rapidly dispersing vorticity. Design~c responds more promptly to the reversal, maintaining chordwise vorticity continuity and stable vortex development along the foil chord. This faster structural response—enabled by the higher loss modulus suppressing residual elastic oscillations—allows the LEV to form earlier in the new half-stroke, directly supporting the larger and more coherent thrust peaks observed in the force measurements.

\section{Unconstrained Rapid Acceleration}

\subsection{Setup}

To validate that the propulsive advantages identified under constrained conditions translate to task-level acceleration performance, unconstrained swimming trials were conducted at $St=0.8$. The CPS $x$-axis was switched from fixed-position mode to a virtual-mass mode that emulates free swimming: the measured streamwise thrust is fed into a real-time dynamic model with a virtual mass of $3$~kg (approximating the mass of a typical robot at comparable scale, the physical mass of the CLD layer is negligible by comparison), and the resulting acceleration is integrated to update the carriage velocity, allowing the foil to freely accelerate along the flow direction. Only the baseline and 
design~c were tested, as these represent the two extremes of the CLD coverage range.

\subsection{Results}

Design~c markedly outperforms the baseline across all kinematic metrics (Fig.~\ref{fig:5}). Over a $3.8$~s trial, design~c achieves a net displacement of $0.617$~m versus $0.213$~m for the baseline, 
with total absolute travel distances of $1.377$~m and $0.973$~m, respectively. The peak acceleration of design~c reaches $0.191~\mathrm{m/s^2}$—nearly five times the baseline value of $0.039~\mathrm{m/s^2}$—and its terminal velocity is approximately three times higher ($0.291$~m/s versus $0.105$~m/s). These 
improvements are consistent with the $>$200\% thrust increase observed in the constrained propulsion tests at the same $St$, confirming that the force-level benefits of CLD-induced impedance shaping transfer directly to free-swimming acceleration.

The trailing-edge kinematics provide further insight. The baseline foil exhibits a distorted, square-wave-like trailing-edge trajectory, indicating that the actuation input overpowers the structural stiffness and excites higher-order bending modes. Design~c, by contrast, maintains a smooth, near-sinusoidal oscillation throughout the trial. This kinematic regularization is consistent with the 
elevated loss modulus acting as a mechanical low-pass filter that suppresses parasitic high-frequency content, preserving the clean waveform needed for coherent vortex formation and efficient thrust 
production.

\section{Discussion and Conclusion}
\subsection{Passive Impedance Shaping as a Design Principle}
The results across all three experimental stages support a consistent physical narrative. The CLD mechanism introduces a frequency-dependent loss modulus that progressively rebalances the impedance composition from elastic-dominant to dissipative-dominant as actuation frequency increases. At the structural level, this manifests as increased energy dissipation per cycle (Section~III). At the fluid--structure interaction level, the elevated dissipation introduces a phase lag that delays the foil's response at stroke reversal, shifting the interaction from abrupt flow separation toward controlled vortex capture (Section~IV). At the task level, the same mechanism acts as a mechanical low-pass filter that suppresses higher-order bending modes, regularizing the trailing-edge kinematics and enabling sustained acceleration (Section~V). This multi-scale consistency also explains the divergent efficiency profiles: whereas the baseline is effectively tuned for a narrow low-$St$ operating point, the CLD-enhanced foil maintains useful efficiency across the full $St$ range by passively adapting its impedance to each regime.

Notably, the phase lag introduced by CLD plays a qualitatively different role than the transmission delays typically penalized in rigid propulsion systems\cite{fossen2011handbook}. In conventional designs, phase lag between actuation and force output represents wasted energy and reduced responsiveness\cite{ogata2020modern}. Here, the CLD-induced delay instead allows the foil to yield smoothly at stroke reversal rather than snapping abruptly against the fluid, promoting stable LEV attachment and coherent vortex shedding\cite{shah2024controlling,feng2025lift}. This reframing—from phase lag as a penalty to phase lag as a hydrodynamic enabler—is central to the proposed strategy, just 
as the low-pass filtering effect of the viscoelastic layer suppresses the parasitic high-frequency content that would otherwise trigger premature flow separation at high $St$.

A notable feature of this approach is that it requires no sensing, computation, or active actuation beyond the primary drive. The impedance adaptation is embedded entirely in the material architecture\cite{laschi2016soft}, making it inherently scalable to multi-degree-of-freedom soft structures where spatially distributed active stiffness control would be impractical\cite{zhong2021tunable,quinn2022tunable}. For practical underwater vehicles, this broad-spectrum capability enables seamless transitions between sustained cruising and burst acceleration without requiring mode-switching or gain scheduling in the controller.

\subsection{Limitations and Future Work}
Several limitations of the current study should be noted. First, as an initial investigation of CLD-based propulsor design, all tests in this study employ a single-degree-of-freedom passive fin; how 
multi-degree-of-freedom CLD systems—whether arranged in series (multi-joint fish-like) or in plane (ray fin-like)—respond to fluid--structure interactions remains an open question. Second, the unconstrained trials use a virtual-mass emulation rather than a physically free-swimming robot, and the one-dimensional carriage dynamics do not capture lateral or rotational degrees of freedom present in real swimming. Third, CLD coverage was tested at only four discrete levels without systematic optimization; the relationships between coverage distribution (spanwise, chordwise), layer orientation, and varying layer thickness on propulsive performance remain unexplored. Consequently, whether a specific CLD-coverage threshold governs the emergence of plateau-like efficiency remains unresolved.

\subsection{Conclusion}

This work demonstrates that constrained-layer damping provides a passive, frequency-selective mechanism for reshaping the impedance composition of soft robotic propulsors. Through structural characterization, constrained propulsion analysis, and unconstrained acceleration trials, we show that CLD shifts the force--velocity phase relationship toward alignment at high Strouhal numbers, yielding a five-fold increase in peak acceleration and a three-fold increase in terminal velocity without active control. These results establish phase-oriented passive impedance modulation as a practical design pathway for enhancing transient propulsion in soft robotic swimmers, and motivate further exploration of spatially graded CLD architectures for multi-modal aquatic locomotion.

\section*{Acknowledgment}
This work was supported by the National Science Foundation 
(NSF Award No.\ 2422534).
\bibliographystyle{IEEEtran}
\bibliography{references}
\nocite{*}

\end{document}